\documentclass[journal]{IEEEtran}

\usepackage{amssymb, amsmath, bm}
\usepackage{multirow, booktabs, boldline}
\usepackage{graphicx, float}
\usepackage[colorlinks=true, pagebackref]{hyperref}
\usepackage{caption}
\usepackage{enumitem}
\graphicspath{{figs/}}

\begin{document}

\title{A Gating Model for Bias Calibration in\\Generalized Zero-shot Learning}

\author{Gukyeong~Kwon, Ghassan AlRegib,~\IEEEmembership{Fellow,~IEEE}
\thanks{Manuscript received February 4, 2021; revised September 5, 2021 and January 4, 2022; accepted
January 30, 2022. G. Kwon and G. AlRegib are with the School
of Electrical and Computer Engineering, Georgia Institute of Technology, Atlanta,
GA, 30332 USA e-mail: gukyeong.kwon@gmail.com, alregib@gatech.edu.}}

\markboth{IEEE Transactions on Image Processing, 2022}%
{Kwon and AlRegib: A Gating Model for Bias Calibration in Generalized Zero-shot Learning}
%



\maketitle

\begin{abstract}
Generalized zero-shot learning (GZSL) aims at training a model that can generalize to unseen class data by only using auxiliary information. One of the main challenges in GZSL is a biased model prediction toward seen classes caused by overfitting on only available seen class data during training. To overcome this issue, we propose a two-stream autoencoder-based gating model for GZSL. Our gating model predicts whether the query data is from seen classes or unseen classes, and utilizes separate seen and unseen experts to predict the class independently from each other. This framework avoids comparing the biased prediction scores for seen classes with the prediction scores for unseen classes. In particular, we measure the distance between visual and attribute representations in the latent space and the cross-reconstruction space of the autoencoder. These distances are utilized as complementary features to characterize unseen classes at different levels of data abstraction. Also, the two-stream autoencoder works as a unified framework for the gating model and the unseen expert, which makes the proposed method computationally efficient. We validate our proposed method in four benchmark image recognition datasets. In comparison with other state-of-the-art methods, we achieve the best harmonic mean accuracy in SUN and AWA2, and the second best in CUB and AWA1. Furthermore, our base model requires at least 20$\%$ less number of model parameters than state-of-the-art methods relying on generative models. 
\end{abstract}


%
\IEEEpeerreviewmaketitle

\section{Introduction}
\IEEEPARstart{A}{dvancement} in machine learning has primarily been driven by a large amount of labeled data. In particular, a supervised learning framework which utilizes fully annotated data such as ImageNet~\cite{deng2009imagenet} achieves state-of-the-art performance in diverse applications such as object recognition, detection, and segmentation~\cite{he2016deep, ren2016faster,chen2017deeplab}. However, supervised learning has clear limitations when generalizing in numerous real-world scenarios because of expensive data collection and annotation. Also, to generalize the supervised model to a new class, the model needs to be trained with a large amount of data for the new class even though the new class is similar to other trained classes. These limitations motivate the development of other learning paradigms that do not require fully annotated data.

Zero-shot learning (ZSL) aims at learning a model that generalizes to untrained classes~\cite{larochelle2008zero, lampert2009learning}. To achieve this goal, auxiliary information such as attributes of the unseen class is utilized. For example, in the application of image recognition, assume that a classifier is trained for `horse' class and `striped cat' class. If we have auxiliary information of textual description for a new class `zebra' such as ``zebra is a horse with stripes'', the classifier can associate the `horse' features and `stripe` features from training images to learn the new class `zebra'. Depending on the evaluation set up, ZSL can be further categorized into the standard ZSL and generalized zero-shot learning (GZSL). In standard ZSL, test images are drawn only from unseen classes. However, GZSL focuses on achieving high accuracy for both seen and unseen class test images. In this paper, we specifically tackle the problem of GZSL for image recognition.

\begin{figure}[t]
\centering
    \includegraphics[width=\linewidth]{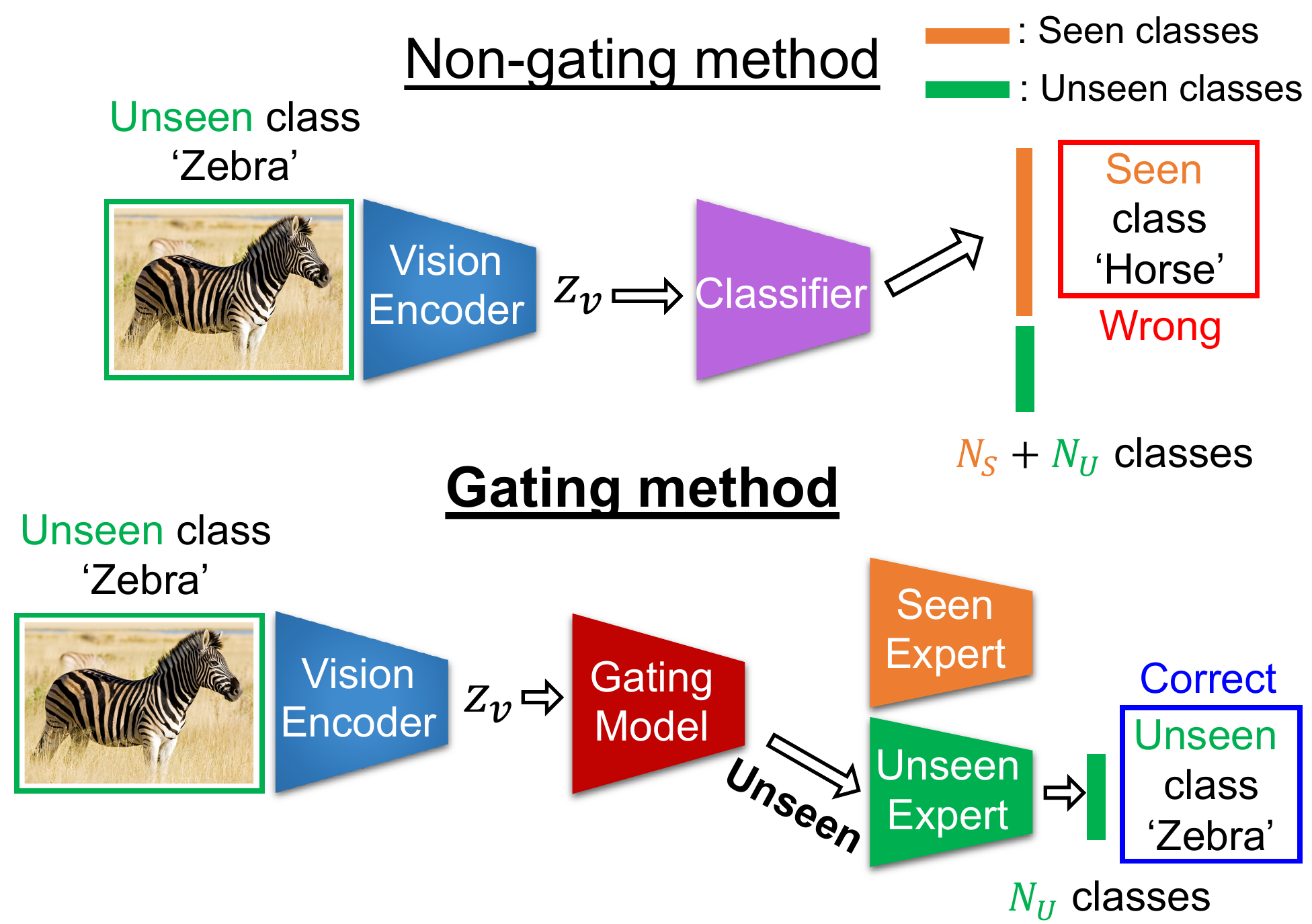}
    \caption{Comparison between the non-gating method and the gating method.}\label{fig:intro}
\end{figure}
One of the main challenges in GZSL is a biased model prediction caused by the inherently unbalanced training set. During the training of GZSL algorithms, both visual and attribute features are available for seen classes while only attribute features are provided for unseen classes. Hence, the unbalanced training set causes models to overfit on seen class data and perform well for seen classes but poorly for unseen classes. Several approaches~\cite{liu2018generalized, chao2016empirical} have been proposed to overcome this challenge by calibrating prediction scores for seen classes. However, we still observe that most of the unseen classes are misclassified as seen classes. In these calibration methods, the classifier makes a prediction out of the search space that contains both seen and unseen classes. Thus, the prediction scores for unseen classes cannot completely avoid competing with the biased prediction scores for seen classes. We propose using a gating model to tackle the biased prediction challenge in GZSL.

In Fig.~\ref{fig:intro}, we compare the standard (non-gating) method and the gating method in GZSL to highlight the differences. For both models, a visual representation, $z_{v}$, is obtained by giving an input image to the vision encoder. Assume that an unseen class image is given to the models. $N_{S}$ and $N_{U}$ denote the number of seen classes and unseen classes. The gating method consists of three components, which are a gating model, a seen expert, and an unseen expert. The seen expert and the unseen expert are trained to correctly classify seen and unseen classes, respectively. The gating model first performs unseen class detection which aims at correctly predicting whether the image is from the seen or the unseen classes. Based on the unseen class detection result, either the seen expert or the unseen expert is chosen to predict the class. While in the standard non-gating method, a class is predicted out of total $N_{S} + N_{U}$ classes, the gating method predicts a class out of either $N_{S}$ classes or $N_{U}$ classes. Thus, the model can avoid comparing the biased seen class prediction scores with the unseen class prediction scores in the gating method. 

We propose a two-stream autoencoder-based gating model which possesses several advantages over other GZSL methods. In particular, we utilize representations from latent space and cross-reconstruction space to characterize association between query visual input and attributes and perform accurate unseen class detection. Also, our two-stream autoencoder provides a unified framework for both the gating model and the unseen expert. The latent representations are trained to be class-discriminant and directly utilized for unseen class classification. Therefore, no additional unseen expert needs to be trained, which leads to the computational efficiency of the proposed method. Furthermore, we show that both experts can be separately optimized and the gating model can be easily combined with other state-of-the-art methods to improve the overall GZSL performance. In summary, the main contributions of this paper are three folds:

\begin{enumerate}[label=\roman*, leftmargin=0.5cm]
    \item We propose a gating model which prevents biased prediction toward seen classes and achieves state-of-the-art performance in four benchmark image recognition datasets.
    \item We validate that the proposed method is easily combined with any existing state-of-the-art methods to further improve the performance. Such generalizability is a contribution by itself to reflect the impact of GatingAE. 
    \item We achieve effective unseen class detection and classification in a unified framework which significantly reduces the number of model parameters.
\end{enumerate}

The structure of this paper is organized as follows. First, we comprehensively review the related works in Section~\ref{sec:related}. Representation learning using the two-stream autoencoder is explained in Section~\ref{sec:representation}. In Section~\ref{sec:unseen}, we discuss the proposed unseen class detection method and classification. The experiments are discussed in Section~\ref{sec:experiments} and we conclude the paper in Section~\ref{sec:conclusion}.
\section{Related Works}\label{sec:related}
We broadly review existing works related to ZSL and GZSL. We categorize the related works into three categories and explain details of the works in each category. 

\textbf{Joint representations for visual features and attributes} Learning joint representations for visual and attribute data is essential to utilize the association between them and solve ZSL problems. In~\cite{DeViSE}, semantic knowledge learned from text data is used as a type of attributes and aligned with visual representations in the joint embedding space. The authors in~\cite{ReViSE} use autoencoders to obtain representations for visual data and textual attributes. Also, a cross-modality distribution matching constraint is imposed to align representations from both modalities. The authors in~\cite{rahman2018unified} propose to learn a projection for each class of images to model the relationship between seen and unseen classes. In~\cite{changpinyo2016synthesized}, the joint representation is learned by matching the graph structure of semantic space and model space. The authors in~\cite{norouzi2013zero} propose to map images to the semantic embedding space through the convex combination of semantic embedding vectors. A latent probabilistic model and a low-rank semantics grouping method are respectively proposed in~\cite{zhang2016zero} and~\cite{xu2021semi} to learn the statistical relationship between visual and attribute representations. The authors in~\cite{LATEM, romera2015embarrassingly} propose to learn compatibility functions that can relate the visual features with attribute representations. In~\cite{TCN}, the authors propose to learn the joint representations through contrastive learning and generalize the representations to unseen classes by imposing a transferability constraint. In~\cite{DAZLE}, a dense attribute-based attention mechanism is proposed to align attributes with local visual features instead of global feature vectors from images. Our two-stream autoencoder shares representations from each stream to learn joint embedding.

\textbf{Generative models for feature generation} Generative models such as generative adversarial networks (GANs)~\cite{goodfellow2014generative} and variational autoencoders (VAEs)~\cite{kingma2013auto} have been widely used to generate unseen class visual features and directly use them for training a classifier. The authors in~\cite{f-CLSWGAN} use a Wasserstein GAN (WGAN)~\cite{arjovsky2017wasserstein} conditioned on attribute information to generate unseen visual features. \cite{felix2018multi} and \cite{GMN} impose a multi-modal cycle consistency loss and gradient matching loss on the WGAN, respectively, to generate class discriminant unseen class data. \cite{long2017zero} uses a diffusion regularization which aims at reducing the reluctant dimensions in the synthesized data and diffusing information to all the dimensions. In~\cite{E-PGN}, a modified WGAN is used to generate visual prototypes in an episode-based training setup. \cite{ABP} uses a single conditional generator trained via an alternating backpropagation algorithm to generate visual features. Instead of GANs, several methods are based on conditional VAEs to synthesize samples~\cite{mishra2018generative, kumar2018generalized}. In~\cite{schonfeld2019generalized}, a two-stream VAE is utilized to generate latent representations for unseen class samples and the latent representations are used to train a classifier. Although generative model-based approaches have achieved successful performance in GZSL, generative models often require a large number of model parameters. Our proposed method is not relying on generative models such as VAEs or GANs and requires significantly less computational resources to train the model. 

\textbf{Calibration of biased prediction toward seen classes} Several works have focused on preventing models from making a biased prediction toward seen classes and achieving high accuracy performance for both seen and unseen classes in GZSL. \cite{socher2013zero} proposes a gating model which estimates the local outlier probability for unseen class detection. The authors in~\cite{zhang2018model} propose a gating model which uses not only a seen and an unseen expert but also a general all class classifiers together to make a prediction. In~\cite{COSMO}, adaptive confidence-based smoothing is utilized with the soft-gating model which combines prediction scores from the seen and the unseen expert. According to this work, our proposed method can be categorized as a hard-gating model which uses either seen or unseen expert for each input. \cite{DVBE} filters out seen class samples by thresholding the entropy of the predicted scores and predicts the seen and the unseen classes separately. The authors in~\cite{liu2018generalized} propose using temperature scaling~\cite{hinton2015distilling} and an entropy-based regularization to mitigate the overfitting on seen class data. \cite{chao2016empirical} calibrates the seen class prediction by using calibrated stacking which reduces the prediction score for seen classes. 

The unseen class detection in the proposed method is largely inspired by techniques in anomaly detection which share the goal of detecting unseen class samples. In particular, an unsupervised learning framework of the autoencoder is widely explored for anomaly detection. Following the classic work~\cite{japkowicz1999concept}, the authors in \cite{sakurada2014anomaly, zhou2017anomaly, sabokrou2018adversarially} use the reconstruction error of the autoencoder to detect anomalies. In~\cite{zong2018deep}, the authors fit Gaussian mixture models (GMMs) to the representations of the autoencoder and the likelihood is measured to detect anomalies. \cite{abati2019latent} detects unseen class samples using an autoregressive density estimation model learned in the latent space of the autoencoder. \cite{kwon2020novelty, kwon2020backpropagated} propose gradient-based representations obtained from the autoencoder to detect anomalies from a model perspective. Compared to the most of the existing anomaly detection algorithms, we characterize unseen class samples using data from two different modalities, vision and attribute. Also, anomaly detection algorithms only need to detect unseen class samples but the proposed algorithm learns to detect and classify seen and unseen class data. Therefore, learning representations that can separate seen and unseen class samples while being discriminant for each class is an critical step in the proposed method.
\section{Representation Learning Using\\ A Two-Stream Autoencoder}\label{sec:representation}
In this section, we define the problem of GZSL and explain the two-stream autoencoder for learning joint representations of visual features and attributes. 

\subsection{Problem Setup}
We first define notations for the training data. Assume that $\mathcal{X}_{tr}^{S}, \mathcal{Y}^S, \mathcal{A}^S$ denote sets of visual features for seen class training images, seen classes, and seen class attributes, respectively. Since there is an associated attribute for each class, the sets for seen classes and their attributes can be written as $\mathcal{Y}^S = \{y_{1}^S, \cdots, y_{|\mathcal{Y}^S|}^S\}$ and $\mathcal{A}^S = \{a_{1}^S, \cdots, a_{|\mathcal{Y}^S|}^S\}$, where $|\mathcal{Y}^S|$ defines the number of seen classes. If the class of a visual feature vector, $x_{tr}^S$, is $y_{i}^{S}$, where $i$ is a class index, the training sample can be given as a pair of $(x_{tr}^S, a_{i}^S, y_{i}^S)$. We also have access to unseen class attributes, $\mathcal{A}^U = \{a_{1}^U, \cdots, a_{|\mathcal{Y}^U|}^U\}$, and their associated unseen classes, $\mathcal{Y}^U = \{y_{1}^U, \cdots, y_{|\mathcal{Y}^U|}^U\}$, but do not have access to unseen class visual features during training. Assume that a set of visual features for seen class test images and that for unseen class test images are denoted as $\mathcal{X}_{te}^S$ and $\mathcal{X}_{te}^U$, respectively. In contrast to the standard ZSL where test images are from unseen classes, a visual feature vector, $x_{te}$, are drawn from the union of seen and unseen class test sets, $\mathcal{X}_{te}^S \cup \mathcal{X}_{te}^U$, in GZSL. The goal of GZSL is to learn a classifier, $f$, which can predict the correct label for $x_{te}$ and it can be formulated as $f(x_{te}, a; \theta_{f}): (\mathcal{X}_{te}^S \cup \mathcal{X}_{te}^U) \times (\mathcal{A}^S \cup \mathcal{A}^U) \rightarrow (\mathcal{Y}^S \cup \mathcal{Y}^U)$, where $\theta_f$ is the model parameters. 

\begin{figure}[t]
  \centering
    \includegraphics[width=0.9\linewidth]{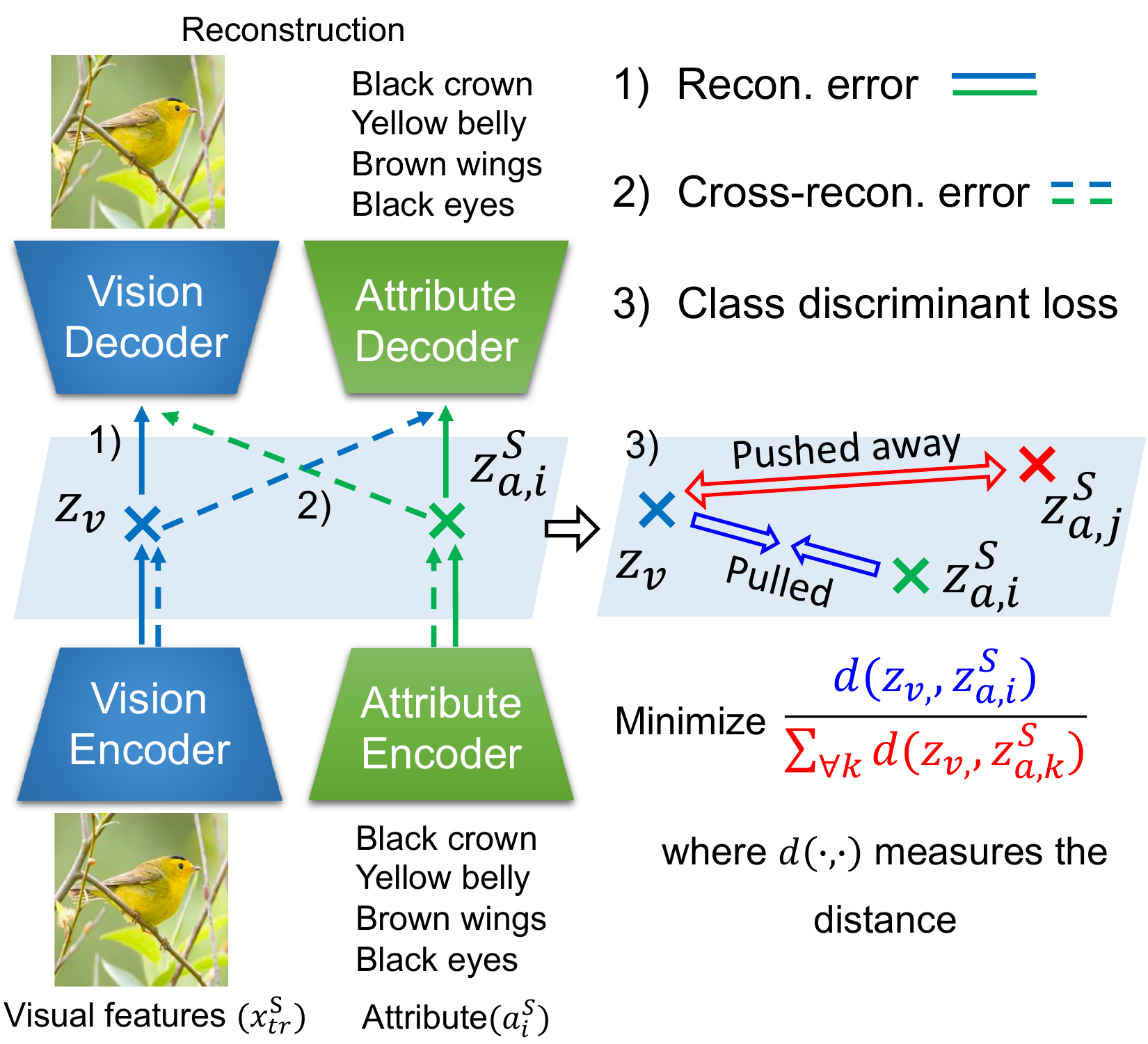}
    \caption{Training of the two-stream autoencoder.}\label{fig:ae_train}
\end{figure}

\subsection{Two-stream Autoencoder}~\label{subsec:autoencoder}
We use a two-stream autoencoder to learn representations that associate visual features with attributes. The two-stream autoencoder consists of a vision stream and an attribute stream. Also, each stream has an encoder and a decoder denoted as $f_{v}$ and $g_{v}$, respectively for the vision stream and $f_{a}$ and $g_{a}$ for the attribute stream. 

We train the autoencoder by imposing three different losses as shown in Fig.~\ref{fig:ae_train}. The first loss is a reconstruction error, $\mathcal{L}_{recon}$. Assume that a vision input, $x_{tr}^{S}$, the class of which is $y_{i}$ and an associated attribute, $a_{i}^{S}$, are given to the autoencoder. Reconstruction for the vision and the attribute input can be denoted as $g_{v}(f_{v}(x_{tr}^{S}))$, $g_{a}(f_{a}(a_{i}^{S}))$, respectively. We measure the $l_1$ distance between the input and the reconstruction for each modality to obtain the reconstruction error. The reconstruction error for each modality is combined as follows:
\begin{equation}
    \mathcal{L}_{recon} = \lVert x_{tr}^{S} - g_{v}(f_{v}(x_{tr}^{S})) \rVert_1 + \lVert a_{i}^{S} - g_{a}(f_{a}(a_{i}^{S})) \rVert_1.
\end{equation}\label{eq:recon}

In addition, we impose a cross-reconstruction error to align representations from visual features and attributes. The cross-reconstruction error has been widely used in the context of multimodal representation learning~\cite{zhao2020multimodal}. In particular, we train the autoencoder model to reconstruct one modality input from the other modality input as depicted in Fig.~\ref{fig:ae_train} 2). The visual features and the attributes are sequentially processed by the vision encoder and the attribute decoder, and attribute encoder and the vision decoder, respectively. We use $l_1$ distance to measure the cross-reconstruction error, $\mathcal{L}_{cross}$, which is formulated as follows:
\begin{equation}
    \mathcal{L}_{cross} = \lVert x_{tr}^{S} - g_{v}(f_{a}(a_{i}^{S})) \rVert_1 + \lVert a_{i}^{S} - g_{a}(f_{v}(x_{tr}^{S})) \rVert_1.
\end{equation}\label{eq:cross} 
We empirically found that the $l_1$ distance for $\mathcal{L}_{recon}$ and $\mathcal{L}_{cross}$ results in better representations for unseen class detection, which will be explained in Section~\ref{sec:unseen}, than $l_2$ distance.

Finally, we train the model with a cross entropy loss, $\mathcal{L}_{cls},$ to obtain class discriminant latent representations. As shown in Fig.~\ref{fig:ae_train} 3), we first obtain the visual latent representation as $z_{v} = f_{v}(x_{tr}^{S})$ and all the seen attribute latent representations as $z_{a, k}^{S} = f_{a}(a_{k}^{S})$, where $k = 1, \dots, |\mathcal{Y}^{S}|$. When the class of visual input is assumed to be $y_i$, the loss is computed as
\begin{equation}
    \mathcal{L}_{cls} = -\log \left(\dfrac{\exp(-\lVert z_{v} - z_{i}^{S}\rVert_2)}{\sum_{k=1}^{|\mathcal{Y}^{S}|}\exp(-\lVert z_{v} - z_{k}^{S}\rVert_2)}\right).
\end{equation}\label{eq:cls}
The term in the numerator contributes to minimize the distance for the positive pair of visual and attribute representations while the terms in the denominator enforce to maximize the distance for negative pairs. 

The overall loss for the autoencoder, $\mathcal{L}_{all}$ is given as follows:
\begin{equation}
    \mathcal{L}_{all} = \mathcal{L}_{recon} + \mathcal{L}_{cross} + \alpha \mathcal{L}_{cls}  ,
\end{equation}
where $\alpha$ is empirically determined to balance the cross entropy loss and the reconstruction losses. The two-stream autoencoder and losses that we use are also commonly explored in other existing works~\cite{schonfeld2019generalized, yu2020episode}. However, we highlight that we still achieve the effective characterization of unseen classes with these generic model and loses. The simplicity of our representation learning framework allows our bias calibration technique based on the unseen class characterization to be easily combined with other existing techniques. 
\begin{figure*}[t]
  \centering
    \includegraphics[width=0.85\linewidth]{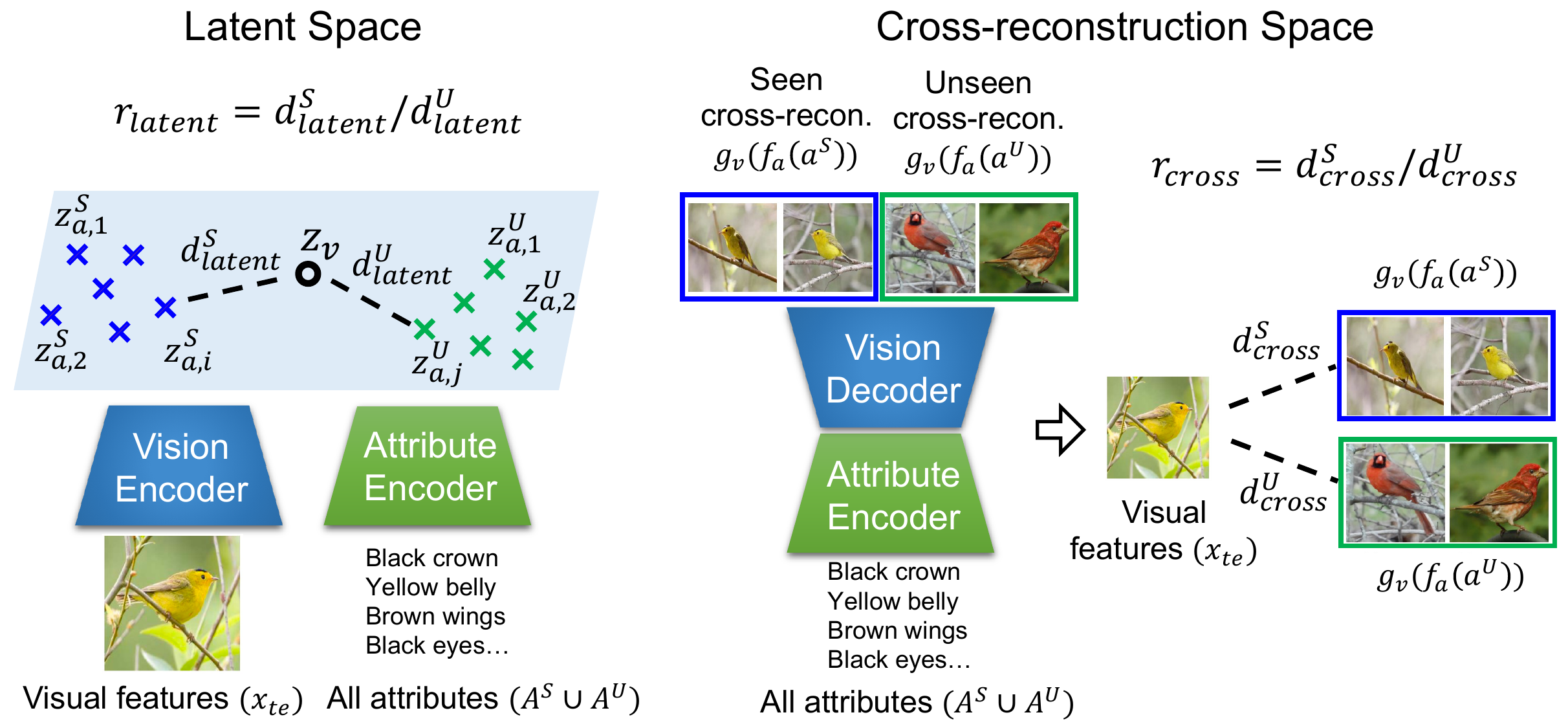}
    \caption{Unseen class detection using distance features in the latent space and the cross-reconstruction space of the two-stream autoencoder.}\label{fig:unseen}
\end{figure*}
\section{Unseen Class Detection and Classification}\label{sec:unseen}
We primarily focus on obtaining descriptive features that can characterize unseen classes from the two-stream autoencoder. In particular, we use the distance between visual and attribute representations as a feature for unseen class detection. The attributes that describe the seen and the unseen classes are available during training and testing. Therefore, we use the attribute representations of the autoencoder as references and compute the distance from the visual representation to the seen and unseen attribute representations. Since the autoencoder is trained to align the seen class visual input and its attribute, the seen class visual input will reside closer to the seen attribute than the unseen class visual input in the representation spaces. The unseen visual representations are not enforced to be aligned with seen attribute nor unseen attribute representations during training. Therefore, from the perspective of the training objectives, the unseen visual representations do not need to reside close to any of seen or unseen attribute representations. However, from the perspective of generalization, since the network has learned to align corresponding visual and attribute representations, we hypothesize that the unseen visual representations are more likely to be aligned with the corresponding unseen attribute representations than seen attribute representations. Hence, by comparing whether the visual representation is closer to the seen or the unseen attribute representations, we can achieve seen and unseen class detection.

We obtain the distance features in both latent space and cross-reconstruction space of the two-stream autoencoder. In particular, the latent space is relatively lower dimension than cross-reconstruction space in our two-stream autoencoder. Therefore, distances obtained in those two spaces abstract features at different semantic levels. We use both low and high dimensional distance features to define the unseen class score which indicates the possibility of the query sample being an unseen class. The detailed steps for unseen class score calculation in both spaces and the final classification are discussed in the following subsections.

\subsection{Unseen class detection in the latent space}
We visualize unseen class detection using the latent representations in the left side of Fig.~\ref{fig:unseen}. The latent representation of the query visual feature, $x_{te}$, is obtained as $z_{v} = f_{v}(x_{te})$. We also generate the latent representations for all the seen and unseen attributes in $\mathcal{A}^{S} \cup \mathcal{A}^{U}$. We denote all the seen and unseen attribute latent representations as $\mathcal{Z}_{a}^{S} = \{z_{a,1}^{S} \dots z_{a,|\mathcal{Y}^S|}^{S}\}$ and $\mathcal{Z}_{a}^{U} = \{z_{a,1}^{U} \dots z_{a,|\mathcal{Y}^U|}^{U}\}$, respectively. We extract distance features by computing the minimum distance from the visual representation to the seen and the unseen attribute representations. The distance features for the seen class, $d_{latent}^{S}$, and the unseen class, $d_{latent}^{U}$, are calculated as follows:
\begin{align}
    d_{latent}^{S} &= \min_{i}\exp (\lVert z_{v} - z_{a, i}^{S}\rVert_{2}) \\
    d_{latent}^{U} &= \min_{j}\exp (\lVert z_{v} - z_{a, j}^{U}\rVert_{2})
\end{align}
We use the exponential of $l_2$ distance since this term is used in the cross entropy loss to align the latent representations during training. We utilize the ratio between these distances to obtain an unseen class score in the latent space, $r_{latent}$, which is defined as $r_{latent} = d_{latent}^{S} / d_{latent}^{U}$. The seen class visual input will result in smaller $d_{latent}^{S}$, larger $d_{latent}^{U}$, and consequently smaller $r_{latent}$ than the unseen class visual input. Therefore, high $r_{latent}$ indicates that the query input is likely to be an unseen class. We can detect the query as an unseen class when the unseen class score is above a certain threshold. Otherwise, the query is detected as a seen class.

\subsection{Unseen class detection in the cross-reconstruction space}
We can also obtain the unseen class score in the cross-reconstruction space as shown in the right side of Fig.~\ref{fig:unseen}. Similar to the calculation of the unseen class score in the latent space, we input all the seen and unseen attributes to the trained attribute encoder. Then, we use the vision decoder to cross-reconstruct images from attributes. The cross-reconstruction of seen and unseen class attributes are denoted as $\{g_{v}(f_{a}(a_{1}^{S})) \dots g_{v}(f_{a}(a_{|\mathcal{Y}^S|}^{S}))\}$ and $\{g_{v}(f_{a}(a_{1}^{U})) \dots g_{v}(f_{a}(a_{|\mathcal{Y}^U|}^{U}))\}$, respectively. We extract the distance features in the cross-reconstruction space by comparing the query visual features and the cross-reconstructions from attributes. The minimum distance from the query visual input to the seen cross-reconstruction, $d_{cross}^{S}$, and to the unseen cross-reconstruction, $d_{cross}^{U}$, are computed as follows:
\begin{align}
    d_{cross}^{S} &= \min_{i} \lVert x_{te} - g_{v}(f_{a}(a_{i}^{S}))\rVert_{1} \\
    d_{cross}^{U} &= \min_{j} \lVert x_{te} - g_{v}(f_{a}(a_{j}^{U}))\rVert_{1}.
\end{align}
$l_{1}$ distance is used by following the cross-reconstruction error imposed during training. We combine two distance features by computing the ratio $r_{cross} = d_{cross}^{S} / d_{cross}^{U}$ and utilize it as an unseen class score from the cross-reconstruction space. When the query visual input is from seen classes, the input should be close to one of seen class cross-reconstructions and achieve smaller $d_{cross}^{S}$ than the unseen class input. Also, from the generalization perspective explained in Section~\ref{sec:unseen}, the unseen class input is more likely to be aligned with one of the unseen cross-reconstructions and result in lower $d_{cross}^{U}$ than the seen class input. Therefore, we can detect unseen classed by comparing $r_{cross}$.

\subsection{Overall unseen class detection and classification}~\label{subsec:overall}
We finally use both distance features obtained in the latent space and the cross-reconstruction space to detect unseen class samples. We obtain the final unseen class score, $r_{all}$, as
\begin{equation}
    r_{all} = \dfrac{\min_{i} \lVert x_{te} - g_{v}(f_{a}(a_{i}^{S}))\rVert_{1} + \beta \exp (\lVert z_{v} - z_{a, i}^{S}\rVert_{2}) }{\min_{j} \lVert x_{te} - g_{v}(f_{a}(a_{j}^{U}))\rVert_{1} + \beta \exp (\lVert z_{v} - z_{a, j}^{U}\rVert_{2})},
\end{equation}
where $\beta$ is a hyperparameter to balance two distances from the latent space and the cross-reconstruction space. We perform baseline experiments to compare the GZSL performance with three different unseen class scores, $r_{latent}$, $r_{cross}$, $r_{all}$. We use $r_{all}$ which shows the best performance in the baseline experiments for the state-of-the-art comparison. For the seen expert, we train a supervised linear classifier with one layer, $f_{cls}^{S}$, using available visual features of seen class training images. For the unseen expert, we do not train any additional model but performs 1-nearest neighbor classification in the latent space to predict the class. To be specific, we measure the $l_2$ distance between visual latent representation and all the unseen attribute latent representations, and the class of the closest unseen attribute representation is predicted as a label. We can formulate the overall seen and unseen class detection, and the classification as follows:
\begin{equation}
\hat{y} = 
\begin{cases}
    f_{cls}^{S}(x_{te}) \quad \text{if} \quad  r_{all} < \tau,\\
    y_{k}^{U} \quad \text{where} \quad k = {\text{arg}\min}_{j} \lVert z_{v} - z_{a, j}^{U}\rVert_{2} \quad \text{else},
\end{cases}
\end{equation}
where $\hat{y}$ is the final class prediction of $x_{te}$. The hyperparameters $\beta$ and $\tau$ are found using the validation set provided in~\cite{xian2018zero}. Also, by following the training protocol described in~\cite{COSMO}, we re-train the model from the scratch using the union of the training and the validation sets after finding $\beta$ and $\tau$. Since we utilize a compact model of the two-stream autoencoder for the gating, the proposed method is called \textbf{\textit{GatingAE}}. 

\section{Experiments}\label{sec:experiments}
We validate the effectiveness of the proposed gating model through rigorous baseline experiments. Also, we highlight the GZSL performance of GatingAE in comparison with other state-of-the-art methods. Finally, comprehensive ablation studies are conducted to experimentally support the advantages of GatingAE.

\begin{figure*}[t]
  \centering
    \includegraphics[width=0.75\linewidth]{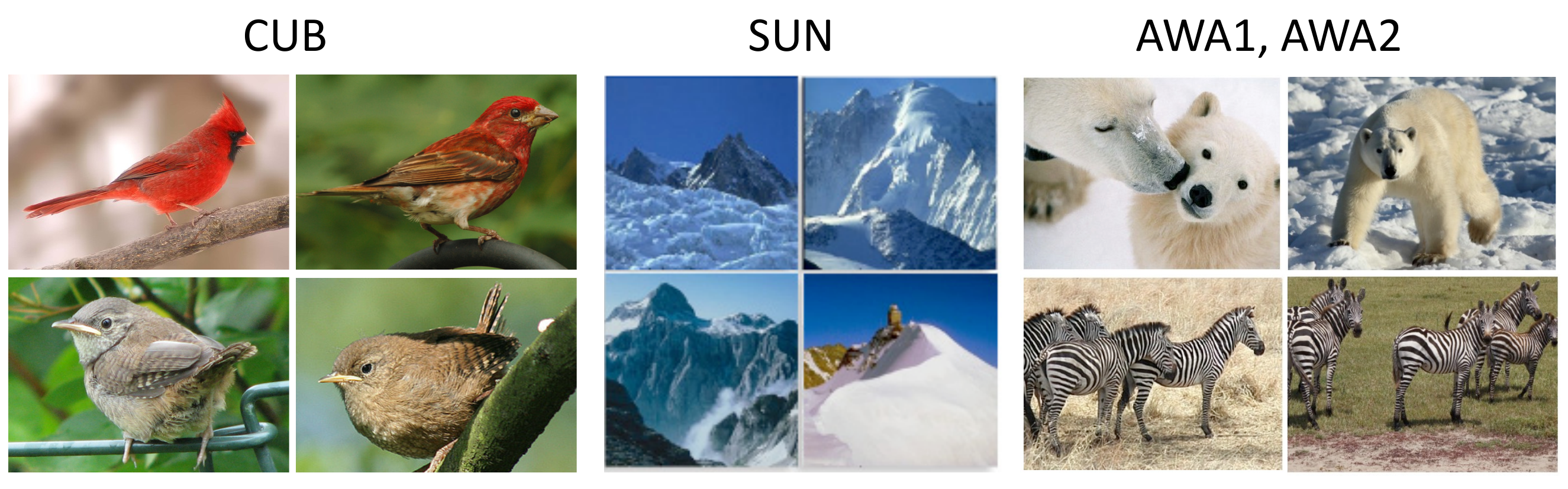}
    \caption{Sample images from CUB, SUN, AWA1, and AWA2.}\label{fig:dataset}
\end{figure*}
\begin{table*}[t]
\centering
\caption{Baseline comparison in CUB, SUN, AWA2, and AWA1 datasets. S: Seen class accuracy, U: Unseen class accuracy, H: Harmonic mean accuracy. Top 2 harmonic mean accuracies for each dataset are highlighed in bold.}\label{tab:baseline}
\begin{tabular}{cccccccccccccc}
\toprule
\multirow{2}{*}{Model} & \multirow{2}{*}{Seen Exeprt} & \multicolumn{3}{c}{CUB} & \multicolumn{3}{c}{SUN} & \multicolumn{3}{c}{AWA2} & \multicolumn{3}{c}{AWA1} \\ \cline{3-14} 
 &  & S & U & H & S & U & H & S & U & H & S & U & H \\ \hline
No gating & 1-NN & 64.4 & 36.6 & 46.8 & 35.0 & 19.0 & 24.7 & 87.8 & 25.9 & 40.0 & 85.3 & 23.8 & 37.2 \\ \hline
\multirow{2}{*}{\begin{tabular}[c]{@{}c@{}}GatingAE\\$(r_{latent})$\end{tabular}} & 1-NN & 55.0 & 54.2 & 54.6 & 29.8 & 48.1 & 36.8 & 81.0 & 54.7 & 65.3 & 76.8 & 54.9 & 64.0 \\ \cline{2-14} 
 & Linear CLF & 58.6 & 54.2 & \textbf{56.4} & 35.7 & 48.1 & \textbf{40.9} & 83.1 & 54.7 & \textbf{66.0} & 78.7 & 54.9 & \textbf{64.7} \\ \hline
\multirow{2}{*}{\begin{tabular}[c]{@{}c@{}}GatingAE\\$(r_{cross})$\end{tabular}} & 1-NN & 45.0 & 58.8 & 51.0 & 27.4 & 50.4 & 35.5 & 79.0 & 55.4 & 65.1 & 73.1 & 55.8 & 63.3 \\ \cline{2-14} 
 & Linear CLF & 47.1 & 58.8 & 52.3 & 32.4 & 50.4 & 39.5 & 80.9 & 55.4 & 65.8 & 74.9 & 55.8 & 63.9 \\ \hline
\begin{tabular}[c]{@{}c@{}}GatingAE $(r_{all})$\end{tabular} & Linear CLF & 58.1 & 54.9 & \textbf{56.4} & 38.1 & 45.4 & \textbf{41.4} & 81.3 & 57.3 & \textbf{67.2} & 72.8 & 59.7 & \textbf{65.6} \\ \bottomrule
\end{tabular}
\end{table*}
\subsection{Experimental Setup}
\textbf{Datasets} We validate the proposed method using four benchmark image recognition datasets: Caltech-UCSD Birds-200-2011 (CUB)~\cite{wah2011caltech}, SUN Attribute (SUN)~\cite{patterson2012sun}, Animals with Attribute 2 (AWA2)~\cite{xian2018zero}, and Animals with Attribute 1 (AWA1)~\cite{lampert2009learning}. Also, we use the proposed splits in ~\cite{xian2018zero} for all the datasets. CUB is a fine-grained dataset with $11,788$ bird images from $200$ species. For the attributes, we use text representations obtained by averaging 10 sentence features per image~\cite{reed2016learning}. SUN is also a fine-grained image dataset which contains $14,340$ visual scene images from $717$ classes. Each scene class is annotated with a $102$-dimensional attribute representation. AWA2 and AWA1 are both coarse-scale image datasets. AWA2 and AWA1 consist of $37,322$ and $30,475$ animal images, respectively. Both datasets have $50$ classes and each class is annotated with a $85$-dimensional attribute representation. As suggested in~\cite{xian2018zero}, we use $2048$-dimensional image representations obtained from the top-layer pooling units of ResNet-101~\cite{he2016deep} pre-trained on ImageNet~\cite{deng2009imagenet} as visual input for all four datasets. Sample images from four datasets are visualized in Fig.~\ref{fig:dataset}.

\textbf{Implementation details}
The encoder and the decoder of each stream in the autoencoder consist of two linear layers and ReLUs are used after the first layer of the encoder and the decoder. The dimension of the latent space is $64$ and the batch size of $64$ is used. The hyperparameter $\alpha$ is searched in the range $[0.01, 0.1]$ with the step size of $0.01$. We use Adam optimizer~\cite{kingma2014adam} with the learning rate of $1.5 \times 10^{-4}$ and train the two-stream autoencoder for $100$ epochs. For the seen expert, we train the one layer linear classifier using the batch size of $32$ and Adam optimizer with the learning rate of $0.001$.

\textbf{Evaluation metrics} We use average per-class top-1 accuracy which is a widely accepted evaluation metric for GZSL to evaluate the proposed method. In particular, we separately calculate the average accuracy for seen classes and unseen classes. We also report the harmonic mean ($H$) of the seen class accuracy ($S$) and the unseen class accuracy ($U$), which is calculated as $H = 2 \times U \times S / (U + S)$. For the evaluation of the unseen class detection performance, we use area under receiver operation characteristic curve (AUC) and false positive rate at true positive rate 0.95 (FPR).

\subsection{Baseline Comparison}\label{subsec:baseline}
We validate the effectiveness of the gating model through comprehensive baseline experiments in Table~\ref{tab:baseline}. We compare the GZSL performance of four different models. All four models are based on the same two-stream autoencoder trained as described in Section~\ref{subsec:autoencoder}. However, the gating approach used in the inference stage differs for them. As shown in the first column of Table~\ref{tab:baseline}, the first model (\textit{No gating}) predicts the class through 1-nearest neighbor (1-NN) classification based on the latent representations of the autoencoder without any gating approaches. To be specific, the predicted class is given as $\hat{y} = y_{k}^{S+U}$, where $k = {\text{arg}\min}_{j} \lVert z_{v} - z_{a, j}^{S+U}\rVert_{2},\: y_{k}^{S+U} \in \mathcal{Y}^S \cup \mathcal{Y}^U,\: z_{j}^{S+U} \in \mathcal{Z}_{a}^{S} \cup \mathcal{Z}_{a}^{U}$. Since no gating approach is used, we note that the classification is made out of $|\mathcal{Y}^{S}| + |\mathcal{Y}^{U}|$ classes. The second and the third models use unseen class scores in the latent space (GatingAE ($r_{latent}$)), and in the cross reconstruction space, (GatingAE ($r_{cross}$)), for gating, respectively. For the two models, we use both 1-NN classifier and a linear classifier (Linear CLF) as seen experts and compare the performance. Finally, we combine distance features obtained in the latent space and the cross-reconstruction space, and use $r_{all}$ as an unseen class score for gating. We use a linear classifier as a seen expert. For all the gating models, 1-NN classifier applied on the latent representations is used as an unseen expert. We report average per-class top-1 accuracy for seen classes (S), unseen classes (U), and the harmonic mean of them (H), in CUB, SUN, AWA2, and AWA1 datasets.

\textbf{Effectiveness of the proposed gating model (No gating vs. GatingAE)} We highlight the contribution of the gating model by comparing the performance of the \textit{No gating} model and GatingAE models using $r_{latent}$ and $r_{cross}$ separately as unseen class scores. For fair comparison, we compare models using the 1-NN classifier as seen experts. GatingAE ($r_{latent}$) with the 1-NN classifier significantly outperforms the \textit{No gating} model by $7.8, 12.1, 25.3,$ and $26.8$ in terms of the harmonic mean accuracy in CUB, SUN, AWA2, and AWA1, respectively. Furthermore, GatingAE ($r_{cross}$) with the 1-NN classifier shows higher harmonic mean accuracy than the No gating model by a margin of $4.2$, $10.8$, $25.1$, and $26.1$ in the four datasets. 

We believe two advantages of GatingAE mainly contribute to the significantly improved performance. First, GatingAE prevents the biased model prediction toward seen classes. The gating model separates the prediction search space and the class is predicted only among seen classes or unseen classes for each sample. This prevents unseen class prediction scores from being directly compared with the biased seen class scores. On the other hand, in the \textit{No gating} model, the biased seen class scores and the unseen class scores are directly compared and the class with the maximum score is predicted. This lead to misclassification of unseen class samples into seen classes. We observe that the seen accuracy of the \textit{No gating} model is at least $1.76$ times and at most $3.58$ times higher than the unseen accuracy of the same model across the four datasets. GatingAE avoids this biased prediction and achieves significantly improved harmonic mean accuracy. Second, the gating approach reduces the dimension of the prediction space for the classifiers. The seen and the unseen experts of GatingAE predict a class out of $|\mathcal{Y}^{S}|$ or $|\mathcal{Y}^{U}|$ classes, respectively, instead of total $|\mathcal{Y}^{S}| + |\mathcal{Y}^{U}|$ number of classes as in the \textit{No gating} model. The reduction of prediction space allows experts to focus on less number
of classes for the classification, which lead to better accuracy
performance. With these two advantages, GatingAE significantly improves the harmonic mean accuracy.

\begin{table*}[t]
\centering
\caption{State-of-the-art comparison in CUB, SUN, AWA2, and AWA1 datasets. S: Seen class accuracy, U: Unseen class accuracy, H: Harmonic mean accuracy. Top 2 harmonic mean accuracies for each dataset are highlighed in bold.}\label{tab:sota}
\begin{tabular}{ccccccccccccc}
\toprule
\multirow{2}{*}{Method} & \multicolumn{3}{c}{CUB} & \multicolumn{3}{c}{SUN} & \multicolumn{3}{c}{AWA2} & \multicolumn{3}{c}{AWA1} \\ \cline{2-13} 
 & S & U & H & S & U & H & S & U & H & S & U & H \\ \hline
LATEM~\cite{LATEM} & 57.3 & 15.2 & 24.0 & 28.8 & 14.7 & 19.5 & 77.3 & 11.5 & 20.0 & 71.7 & 7.3 & 13.3 \\ \hline
DeViSE~\cite{DeViSE} & 53.0 & 23.8 & 32.8 & 27.4 & 16.9 & 20.9 & 74.7 & 17.1 & 27.8 & 68.7 & 13.4 & 22.4 \\ \hline
f-CLSWGAN~\cite{f-CLSWGAN} & 57.7 & 43.7 & 49.7 & 36.6 & 42.6 & 39.4 & 68.9 & 52.1 & 59.4 & 61.4 & 57.9 & 59.6 \\ \hline
ReViSE~\cite{ReViSE} & 28.3 & 37.6 & 32.3 & 20.1 & 24.3 & 22.0 & 39.7 & 46.4 & 42.8 & 37.1 & 46.1 & 41.1 \\ \hline
CADA-VAE~\cite{schonfeld2019generalized} & 53.5 & 51.6 & 52.4 & 35.7 & 47.2 & 40.6 & 75.0 & 55.8 & 63.9 & 72.8 & 57.3 & 64.1 \\ \hline
TCN~\cite{TCN} & 52.0 & 52.6 & 52.3 & 37.3 & 31.2 & 34.0 & 65.8 & 61.2 & 63.4 & 76.5 & 49.4 & 60.0 \\ \hline
ABP~\cite{ABP} & 54.8 & 47.0 & 50.6 & 36.8 & 45.3 & 40.6 & 72.6 & 55.3 & 62.6 & 67.1 & 57.3 & 61.8 \\ \hline
COSMO~\cite{COSMO} & 57.8 & 44.4 & 50.2 & 37.7 & 44.9 & 41.0 & - & - & - & 80.0 & 52.8 & 63.6 \\ \hline
GMN~\cite{GMN} & 54.3 & 56.1 & 55.2 & 33.0 & 53.2 & 40.7 & - & - & - & 71.3 & 61.1 & 65.8 \\ \hline
E-PGN~\cite{E-PGN} & 61.1 & 52.0 & 56.2 & - & - & - & 83.5 & 52.6 & 64.6 & 83.4 & 62.1 & \textbf{71.2} \\ \hline
3ME~\cite{3ME} & 60.1 & 49.6 & 54.3 & - & - & - & - & - & - & 65.7 & 55.5 & 60.2 \\ \hline
DAZLE~\cite{DAZLE} & 59.6 & 56.7 & \textbf{58.1} & 24.3 & 52.3 & 33.2 & 75.7 & 60.3 & 67.1 & - & - & - \\ \hline
DVBE*~\cite{DVBE} & 60.2 & 53.2 & 56.5 & 37.2 & 45.0 & 40.7 & 70.8 & 63.6 & 67.0 & - & - & - \\ \clineB{1-13}{2}
\textbf{GatingAE} & 58.1 & 54.9 & 56.4 & 38.1 & 45.4 & \textbf{41.4} & 81.3 & 57.3 & \textbf{67.2} & 72.8 & 59.7 & 65.6 \\ \hline
\textbf{GatingAE + f-CLSWGAN} & 58.1 & 55.4 & \textbf{56.7} & 38.1 & 45.3 & \textbf{41.4} & 81.3 & 60.3 & \textbf{69.3} & 72.3 & 62.5 & \textbf{67.2} \\ \bottomrule
\end{tabular}
\end{table*}
\begin{figure*}[t]
  \centering
    \includegraphics[width=0.9\linewidth]{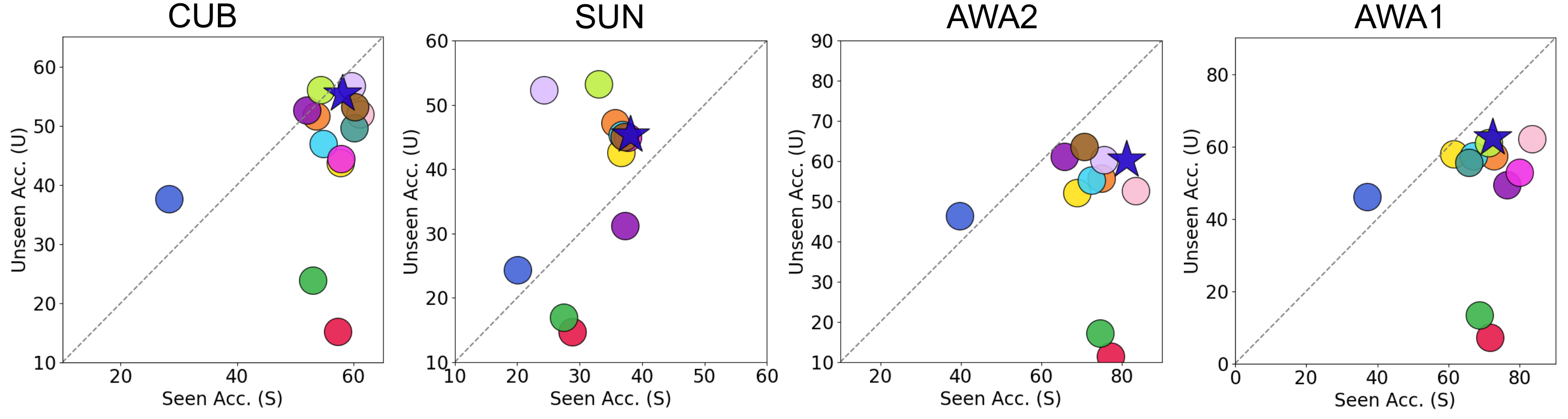}\vspace{-0.5cm}
\end{figure*}
\begin{figure*}[t]
  \centering
  \includegraphics[width=0.7\linewidth]{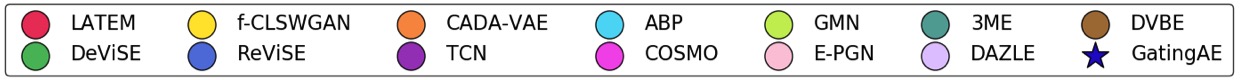}
  \caption{Scatter plot of seen and unseen accuracy for each state-of-the-art algorithm. For an ideal GZSL algorithm, the data point is expected to stay close the middle gray dotted line and the top right corner.}\label{fig:scatter}
\end{figure*}
\textbf{Advantage of using an independently trained expert (1-NN vs. Linear CLF)} We compare the performance of GatingAEs using the 1-NN classifier and the linear classifier (Linear CLF) as seen experts. By comparing these two models, we emphasize that GatingAE can be easily combined with any independently trained expert to further improve the performance. As a case study, we train a linear classifier as a seen expert independently from the gating model or the unseen expert using available visual training data. We show that the linear classifier can improve the seen accuracy without sacrificing unseen class accuracy. In Table~\ref{tab:baseline}, GatingAE ($r_{latent}$) and GatingAE ($r_{cross}$) are combined with the independently trained linear classifiers to achieve higher seen class accuracy than the GatingAE with the 1-NN classifier by at least $1.9$ and $1.8$, respectively, in all four datasets while not compromising the unseen class accuracy. This highlights the applicability of GatingAE with any contributions from seen or unseen experts to further improve the GZSL performance. 

\textbf{Complementary distance features for gating (GatingAE ($\bm{r_{latent}, r_{cross}}$) vs. GatingAE ($\bm{r_{all}}$))} 
We compare GatingAE ($r_{latent}$) and GatingAE ($r_{cross}$) with GatingAE ($r_{all}$) to show the contribution of descriptive distance features from both latent space and the cross-reconstruction space on the GZSL performance. In particular, we compare GatingAEs using the linear classifiers as seen experts because they achieve better performance than GatingAEs using the 1-NN classifiers. In Table~\ref{tab:baseline}, GatingAE ($r_{all}$) consistently achieves higher harmonic mean accuracy than GatingAE ($r_{latent}$) and GatingAE ($r_{cross}$) across all the datasets except that GatingAE ($r_{all}$) achieves the same harmonic mean accuracy as GatingAE ($r_{cross}$) in CUB. We believe the better performance of GatingAE ($r_{all}$) is resulted from the complementary distance features obtained in the latent space and the cross-reconstruction space. Considering that the latent space is lower dimensional than the cross-reconstruction space, the distance features from different spaces contribute to perform gating at different levels of data abstraction. Therefore, by combining both features for $r_{all}$, GatingAE ($r_{all}$) utilizes the advantages of each feature and achieves higher harmonic mean accuracy than both GatingAE ($r_{latent}$) and GatingAE ($r_{cross}$).

\subsection{Comparison With State-of-the-art Algorithms}\label{subsec:sota}
We compare GatingAE with 13 state-of-the-art GZSL algorithms and report the performance in Table~\ref{tab:sota}. A hyphen (-) indicates that the authors of the algorithm have not validated their method in the corresponding dataset. For fair comparison with DVBE, we use their reported performance without finetuning the backbone architecture of ResNet-101 for visual feature extraction. Excluding GatingAE + f-CLSWGAN, the base GatingAE achieves the best harmonic mean accuracy in SUN and AWA2, and the third highest harmonic mean accuracy in CUB and AWA1. Although GatingAE does not achieves the best performance in CUB and AWA1, GatingAE performs more robustly across datasets compared to other algorithms. For instance, DAZLE achieves the highest harmonic mean accuracy in CUB but its harmonic mean accuracy in SUN is $9^{th}$ highest out of 14 algorithms. Also, although E-PGN achieves the highest harmonic mean accuracy in AWA1, its harmonic mean accuracies in CUB and AWA2 are both $4^{th}$ highest out of 14. GatingAE achieves the highest average rank of $2$ over all four datasets in terms of the harmonic mean accuracy. In comparison with the state-of-the-art soft-gating model COSMO, GatingAE which is based on the hard-gating achieves better performance in all datasets. Since the soft-gating model predicts a class using the combination of seen and unseen class prediction scores, the bias toward seen classes still affects the classification of unseen classes. However, GatingAE completely separates the classification of seen and unseen classes and mitigates the effect of the bias in unseen class classification.

We visualize the seen and the unseen accuracies of all the state-of-the-art methods in Fig~\ref{fig:scatter} to analyze the balance between the seen and the unseen accuracies. In particular, the x-axis and the y-axis in each scatter plot indicate the seen accuracy and the unseen accuracy of each method, respectively. Also, the gray dotted line in the middle indicates the same seen and unseen accuracies. An ideal GZSL method should achieve high accuracy for both seen and unseen classes and should not be biased toward either seen classes or unseen classes. Therefore, the accuracy of the ideal method is expected to be plotted close to the top right corner while staying close the dotted gray line. In CUB, GatingAE is one of the most closest methods to the dotted gray line and the top right corner. While DAZLE and GMN are located close to GatingAE in CUB, they are biased toward the unseen class accuracy and located far away from the center line in SUN. Although there are several methods staying close to the center line than GatingAE in AWA2, GatingAE still achieves the highest harmonic mean acccuracy in AWA2. In AWA1, GatingAE shows comparable performance to E-PGN and GMN while being located close to the center dotted line. This shows that GatingAE achieves generalized high accuracy performance for both seen and unseen classes across all four datasets.

We also show that GatingAE can be easily combined with other state-of-the art methods to further improve the performance. Since each expert can be independently improved in GatingAE, the state-of-the-art methods can be simply utilized as a seen or an unseen expert. Furthermore, GatingAE can benefit from the state-of-the-art methods based on generative models, although the state-of-the-art methods do not achieve better GZSL performance than GatingAE. As a case study, we use f-CLSWGAN which is one of the earliest GZSL methods based on a WGAN~\cite{arjovsky2017wasserstein}. f-CLSWGAN generates unseen visual features to tackle the problem of GZSL. We use these generated unseen visual features from f-CLSWGAN to finetune and improve the unseen expert independently from the seen expert. We report the performance of GatingAE + f-CLSWGAN in Table~\ref{tab:sota}. The base GatingAE significantly outperforms f-CLSWGAN by a margin of $6.7$, $2.0$, $7.8$, and $6.0$ in CUB, SUN, AWA2, and AWA1, respectively. However, GatingAE still benefits from f-CLSWGAN and GatingAE + f-CLSWGAN achieves higher harmonic mean accuracy than individual GatingAE and f-CLSWGAN. Since we only finetune the unseen expert, GatingAE + f-CLSWGAN improves the unseen class accuracy over GatingAE while keeping the seen accuracy intact. Also, GatingAE + f-CLSWGAN achieves the best performance in SUN and AWA2, and the second best performance in CUB and AWA1 in terms of the harmonic mean accuracy. Although we only show one case study of using f-CLSWGAN, the same finetuning approach can be utilized with other GZSL algorithms based on generative models such as~\cite{felix2018multi, GMN, mishra2018generative}. 

\begin{table*}[t]
\centering
\caption{Gating performance comparison between GatingAEs and gating models proposed in COSMO~\cite{COSMO}. Ideally, higher harmonic mean accuracy (H), higher AUC, and lower false positive rate at true positive rate 0.95 (FPR) are desired. Top 2 scores in each evaluation metric are highlighted.}\label{tab:gating}
\begin{tabular}{cccccccccc}
\toprule
\multirow{2}{*}{Gating Model} & \multicolumn{3}{c}{CUB} & \multicolumn{3}{c}{SUN} & \multicolumn{3}{c}{AWA1} \\ \cline{2-10} 
 & H$(\uparrow)$ & AUC$(\uparrow)$ & FPR$(\downarrow)$ & H$(\uparrow)$ & AUC$(\uparrow)$ & FPR$(\downarrow)$ & H$(\uparrow)$ & AUC$(\uparrow)$ & FPR$(\downarrow)$ \\ \hline
MAX-SOFTMAX-3~\cite{COSMO} & 43.6 & 0.734 & 0.796 & 38.4 & 0.610 & 0.923 & 53.1 & 0.886 & 0.568 \\ \hline
CB-GATING-3~\cite{COSMO} & 44.7 & 0.820 & 0.720 & \textbf{40.1} & \textbf{0.777} & \textbf{0.775} & 56.8 & \textbf{0.925} & \textbf{0.455} \\ \hline
\begin{tabular}[c]{@{}c@{}}GatingAE $(r_{latent})$\end{tabular} & \textbf{75.7} & \textbf{0.972} & \textbf{0.143} & 38.0 & \textbf{0.777} & \textbf{0.775} & \textbf{62.1} & 0.889 & 0.566 \\ \hline
\begin{tabular}[c]{@{}c@{}}GatingAE $(r_{cross})$\end{tabular} & 61.7 & 0.926 & 0.324 & 34.8 & 0.753 & 0.820 & 61.3 & 0.890 & 0.561 \\ \hline
\begin{tabular}[c]{@{}c@{}}GatingAE $(r_{all})$ \end{tabular} & \textbf{74.9} & \textbf{0.970} & \textbf{0.156} & \textbf{38.8} & \textbf{0.779} & \textbf{0.762} & \textbf{62.7} & \textbf{0.894} & \textbf{0.550} \\ \bottomrule
\end{tabular}
\end{table*}
\subsection{Ablation Study}\label{subsec:ablation}
\textbf{Gating performance comparison with COSMO} We compare the gating performance of GatingAE with the state-of-the-art gating method COSMO~\cite{COSMO} in Table~\ref{tab:gating}. In particular, the authors of COSMO split the validation set into a Gating-Train set and a Gating-Val set and report the gating performance in the Gating-Val set. Following the same protocol, we also train the two-stream autoencoder in the original training set, tune the hyperparameters in the Gating-Train set, and finally report the gating performance in the Gating-Val set. We compare GatingAEs based on $r_{latent}$, $r_{cross}$, $r_{all}$ with two gating models proposed in COSMO, which are MAX-SOFTMAX-3 and GB-GATING-3. The gating performance is validated in terms of the harmonic mean accuracy, AUC, and FPR. We note that GatingAE achieves higher harmonic mean accuracy than COSMO in all the test sets of CUB, SUN, and AWA1 as shown in Table~\ref{tab:sota}. 

GatingAEs significantly outperform MAX-SOFTMAX-3 and GB-GATING-3 in terms of all evaluation metrics in CUB. This further supports the significant performance gap between GatingAE and COSMO in the test set of CUB shown in Table~\ref{tab:sota}. In SUN, while GatingAE ($r_{all}$) achieves slightly lower harmonic mean accuracy in the Gating-Val set compared to CB-GATING-3, it achieves better detection performance with higher AUC and lower FPR. In AWA1, GatingAE ($r_{latent}$) and GatingAE ($r_{all}$) achieve significantly higher harmonic mean accuracy  while achieving lower AUC and higher FPR than CB-GATING-3 in the GZSL-val set. As shown in Table~\ref{tab:sota}, GatingAE ($r_{all}$) outperforms COSMO by a large margin of 2.0 harmonic mean accuracy in the test set of AWA1. Considering that the GZSL-Val set is around three time smaller than the test set of AWA1, we argue that GatingAE ($r_{all}$) maintains its gating performance and learns better class discriminant representations in the relatively large-scale test set of AWA1.

\textbf{Analysis on the gating performance from each distance feature} We decompose the unseen class scores used in GatingAE to understand the contribution of each distance feature on gating. In particular, we report the AUC scores obtained by separately using the latent space distance features, $d_{latent}^{S}$ and $1/d_{latent}^{U}$, and the cross-reconstruction space distance features, $d_{cross}^{S}$ and $1/d_{cross}^{U}$, as unseen class scores. Also, we compare the AUC scores from individual distance features with those from $r_{latent}$, $r_{cross}$, and $r_{all}$, which are the combination of the distance features. This highlights that the distance features are complementary to each other for gating. We report the AUC scores obtained in the test set of CUB, SUN, AWA2, and AWA1 in Table~\ref{tab:auc}. $d_{latent}^{S}/d_{latent}^{U}$ and  $d_{cross}^{S}/d_{cross}^{U}$ shows significant improvement of the AUC scores over $d_{latent}^{S}$, $1/d_{latent}^{U}$, $d_{cross}^{S}$, $1/d_{cross}^{U}$. This shows that the distance features from seen and unseen classes are combined to effectively classify whether the query is from seen classes or unseen classes. In addition, $r_{all}$ shows higher AUC than $r_{latent}$ and $r_{cross}$ in SUN, AWA2, and AWA1. In CUB, $r_{latent}$ performs marginally better than $r_{all}$. We argue that GatingAE combines all the complementary distance features from seen classes, unseen classes, latent space, and cross-reconstruction space to achieve accurate gating results, and consequently choose correct experts for tackling GZSL problems. 
\begin{table}[t]
\centering
\caption{AUC performance obtained from using the distance in the latent space and the cross-reconstruction space as an unseen class score.}\label{tab:auc}
\begin{tabular}{ccccc}
\toprule
Unseen Class Score & CUB & SUN & AWA2 & AWA1 \\ \hline\\[-1em]
$d_{latent}^{S}$ & 0.511 & 0.546 & 0.686 & 0.650 \\\\[-1em] \hline \\[-1em]
$1 / d_{latent}^{U}$ & 0.596 & 0.520 & 0.459 & 0.516 \\\\[-1em] \hline \\[-1em]
$d_{latent}^{S} / d_{latent}^{U} (r_{latent})$ & \textbf{0.842} & \textbf{0.774} & \textbf{0.934} & \textbf{0.917} \\\\[-1em] \clineB{1-5}{2} \\[-1em]
$d_{cross}^{S}$ & 0.496 & 0.550 & 0.725 & 0.697 \\\\[-1em] \hline \\[-1em]
$1 / d_{cross}^{U}$ & 0.574 & 0.500 & 0.421 & 0.427 \\\\[-1em] \hline \\[-1em]
$d_{cross}^{S} / d_{cross}^{U} (r_{cross})$ & \textbf{0.808} & \textbf{0.769} & \textbf{0.933} & \textbf{0.907} \\\\[-1em] \clineB{1-5}{2} \\[-1em]
$r_{cross} + \beta r_{latent} (r_{all})$ & \textbf{0.841} & \textbf{0.783} & \textbf{0.940} & \textbf{0.918} \\ \bottomrule
\end{tabular}
\end{table}

\begin{figure*}[t]
  \centering
    \includegraphics[width=0.75\linewidth]{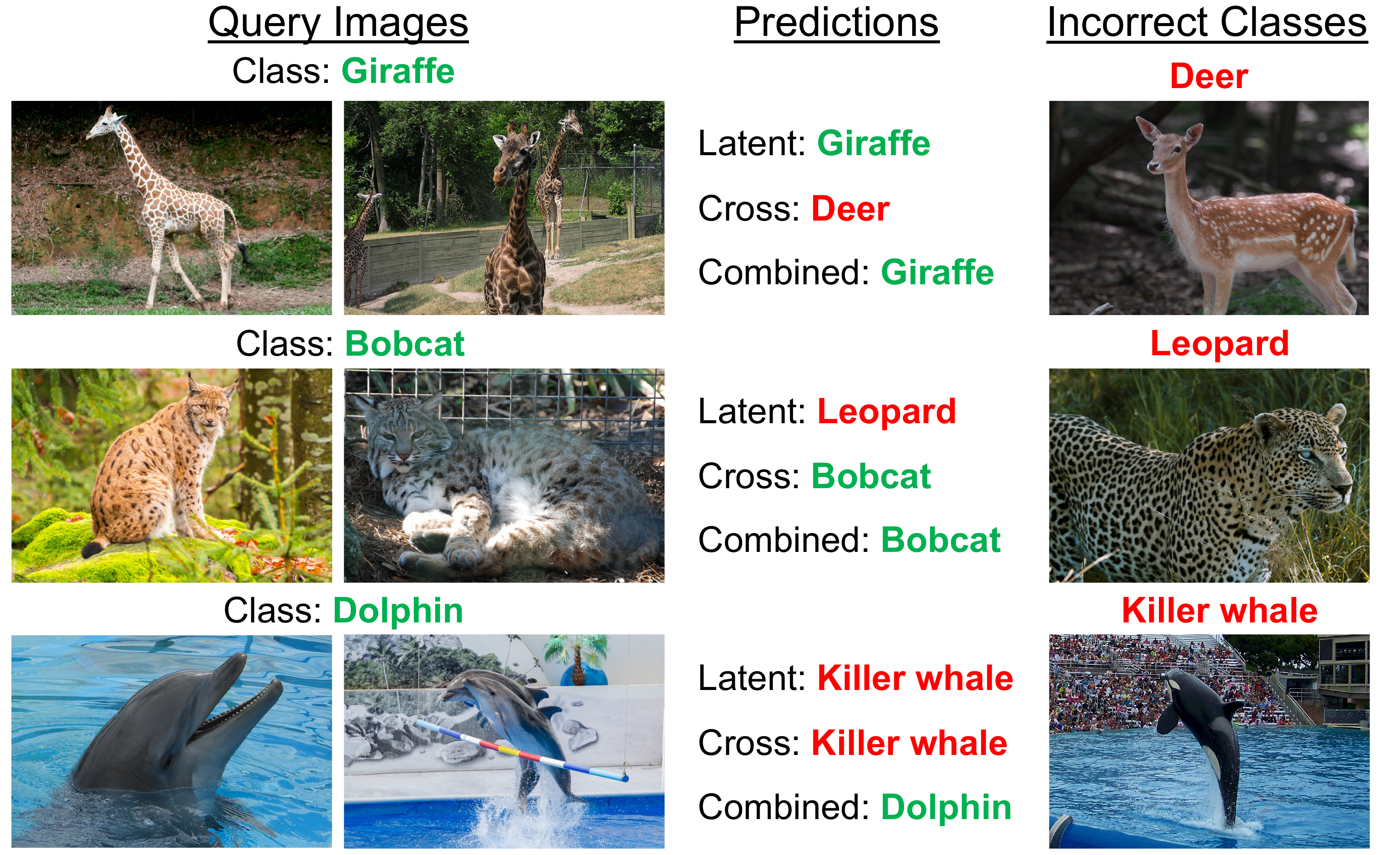}
    \caption{Qualitative analysis on the failure cases of GatingAEs using unseen class scores from different representation spaces in AWA2. Latent, Cross, and Combined refer to the class predictions of GatingAEs using $r_{latent}$, $r_{cross}$, and $r_{all}$, respectively.}\label{fig:failure}
\end{figure*}
\begin{table}[t]
\centering
\caption{Comparison of the number of model parameters between GatingAE and other generative model-based GZSL algorithms.}\label{tab:param}
\begin{tabular}{cccc}
\toprule
Model & f-CLSWGAN~\cite{f-CLSWGAN} & CADA-VAE~\cite{schonfeld2019generalized} & GatingAE \\ \hline
\# of parameters & 19,514,062 & 7,398,716 & 5,860,138 \\ \bottomrule
\end{tabular}
\end{table}

\textbf{Qualitative analysis on complementary distance features} We perform qualitative analysis on the failure cases of GatingAEs using $r_{latent}$, $r_{cross}$, and $r_{all}$ in Figure~\ref{fig:failure}. In particular, unseen class query images are given to GatingAEs and we analyze cases where either GatingAE ($r_{latent}$) or GatingAE ($r_{cross})$ fails, and both of them fail in predicting correct classes. Through this analysis, we further highlight that the latent and the cross-reconstruction features capture unseen classes at different levels of data abstraction. The distance features from the low dimensional latent space focus more on abstracted global features while the cross-reconstruction features capture low-level local characteristics. In the first row of Figure~\ref{fig:failure}, only GatingAE ($r_{cross}$) misclassfies the unseen class \texttt{Giraffe} into the seen class \texttt{Deer}. \texttt{Giraffe} and \texttt{Deer} shares local features of brown and white furs. However, they are clearly distinguished by the global features of the \texttt{Giraffe} such as a long neck and long legs. The unseen class score $r_{latent}$ captures these global features that $r_{cross}$ misses to predict the correct class. In the second row, \texttt{Bobcat} is misclassified as \texttt{Leopard} by GatingAE ($r_{latent}$) while being correctly classified by GatingAE ($r_{cross}$). Since both classes are in the cat family, they are differentiated only by low-level local features such as sharpness of the ears and body patterns. We believe these local features are better captured by $r_{cross}$ than $r_{latent}$. Finally, in the last row, we show two examples of the \texttt{Dolphin} class where both GatingAE ($r_{latent}$) and GatingAE ($r_{cross}$) misclassify them into the seen class \texttt{Killer whale} while GatingAE ($r_{all}$) correctly predicts the unseen class. \texttt{Dolphin} shares most of the characteristic features with and \texttt{Killer whale}, which makes unseen class detection challenging. However, we incorporate both local and global features from $r_{latent}$ and $r_{cross}$ in GatingAE ($r_{all}$) and achieve the correct prediction. For all the query images given in Figure~\ref{fig:failure}, GatingAE ($r_{all}$) predicts the correct classes when one or both of GatingAE ($r_{latent}$) and GatingAE ($r_{cross}$) fail. This show that GatingAE ($r_{all}$) effectively combines the advantages of each feature abstracted at different levels.

\textbf{Computational efficiency of GatingAE} GatingAE is computationally efficient because of its compact two-stream linear autoencoder and the unified framework for the gating model and the unseen expert. To highlight the computational efficiency of GatingAE, we compare the number of parameters required to be trained for GatingAE with f-CLSWGAN and CADA-VAE in Table~\ref{tab:param}. Several state-of-the-art methods such as~\cite{felix2018multi, GMN} are developed on top of f-CLSWGAN. Also, CADA-VAE uses a two-stream VAE which is the closest architecture to our two-stream linear autoencoder. By comparing with these two models which are based on the simple generative models, we emphasize that GatingAE is even simpler while achieving the state-of-the-art performance. As shown in Table~\ref{tab:param}, f-CLSWGAN and CADA-VAE require around $3.3$ times and $1.3$ times more parameters than GatingAE, respectively. CADA-VAE uses the same number of layers and the same dimension for the latent space as GatingAE. However, CADA-VAE has to learn more parameters for a latent constraint and a classifiers for $|\mathcal{Y}^{S}| + |\mathcal{Y}^{U}|$ classes while GatingAE only needs to train a classifier for $|\mathcal{Y}^{S}|$ number of seen classes. In addition, the state-of-the art soft-gating model COSMO uses f-CLSWGAN as an unseen expert. Hence, COSMO requires to train more than 19 million parameters of f-CLSWGAN. However, GatingAE uses the 1-NN classifier which does not need to train any additional parameters as an unseen expert. Therefore, GatingAE uses significantly less computational resources while outperforming these state-of-the-art methods. 

We also compare the computational time for training one epoch of GatingAE and f-CLSWGAN using a single GPU GeForce GTX TITAN X. For f-CLSWGAN, we use the official code released by the authors to measure the training time. We use AWA2 and AWA1 datasets for this experiment since these datasets have about twice to three times more training samples that CUB and SUN. In both AWA2 and AWA1 datasets, GatingAE takes $3.370$ seconds and $2.893$ seconds for training one epoch, respectively, while f-CLSWGAN takes $5.501$ seconds and $4.623$ seconds, respectively. GatingAE requires only about $61 \%$ of training time that is required by f-CLSWGAN. This is mainly because that f-CLSWGAN is based on a GAN which is adversarially trained with more model parameters of the generator and the discriminator than GatingAE. Given that f-CLSWGAN is based on the simplest GAN for feature generating in GZSL, we believe GatingAE has a strong advantage in computational efficiency compared to other state-of-the-art methods using more complicated GAN-based models~\cite{felix2018multi, GMN}. 
\section{Conclusion}\label{sec:conclusion}
We propose a GZSL algorithm, GatingAE, which utilizes the two-stream autoencoder as a gating model to prevent biased prediction and achieve high accuracy performance for both seen and unseen classes data. In particular, we utilize distance features obtained from the latent space and the cross-reconstruction space of the autoencoder for gating. Based on the gating results, either the seen or the unseen class expert is chosen to perform the target task. We thoroughly validate the gating performance and the overall GZSL performance in the application of image recognition. GatingAE achieves the state-of-art performance in four benchmark image recognition datasets. Also, several advantages of GatingAE such as complementary distance features for gating, using independently trained experts, and computational efficiency are highlighted through baseline experiments and ablation studies. We plan to further explore the characterization of the bias presented in the training data and the utilization of the bias information to calibrate the prediction for learning with limited data.

\bibliographystyle{IEEEtran}
\bibliography{reference}
%

\begin{IEEEbiography}[{\includegraphics[width=1in,height=1.25in,clip,keepaspectratio]{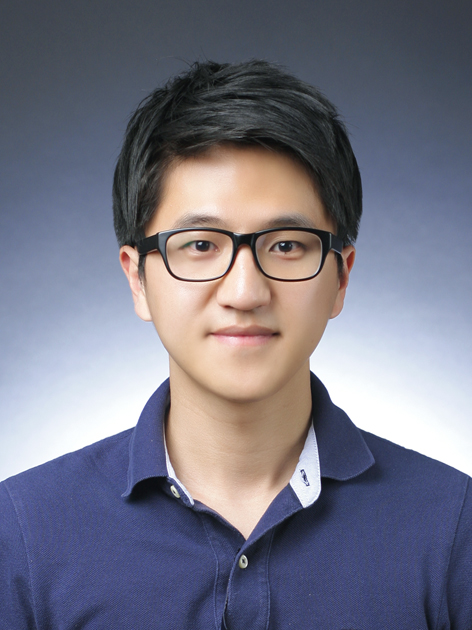}}]{Gukyeong Kwon} is currently an Applied Scientist at Amazon Web Services AI Labs. He received his M.S. and Ph.D. degrees in the School of Electrical and Computer Engineering (ECE) at  Georgia Institute of Technology in 2018 and 2021, respectively. He is a co-recipient of the Finalist of the World's First 10K Best Paper Award at the IEEE International Conference on Multimedia and Expo in 2017, the Best Paper Award at the IEEE International Conference on Image Processing in 2019, and the Top Viewed Special Session Paper Award at the IEEE International Conference on Image Processing in 2020. His research has primarily focused on the robustness of machine learning, multimodal representation learning, and learning with limited data.
\end{IEEEbiography}

\begin{IEEEbiography}[{\includegraphics[width=1in,height=1.25in,clip,keepaspectratio]{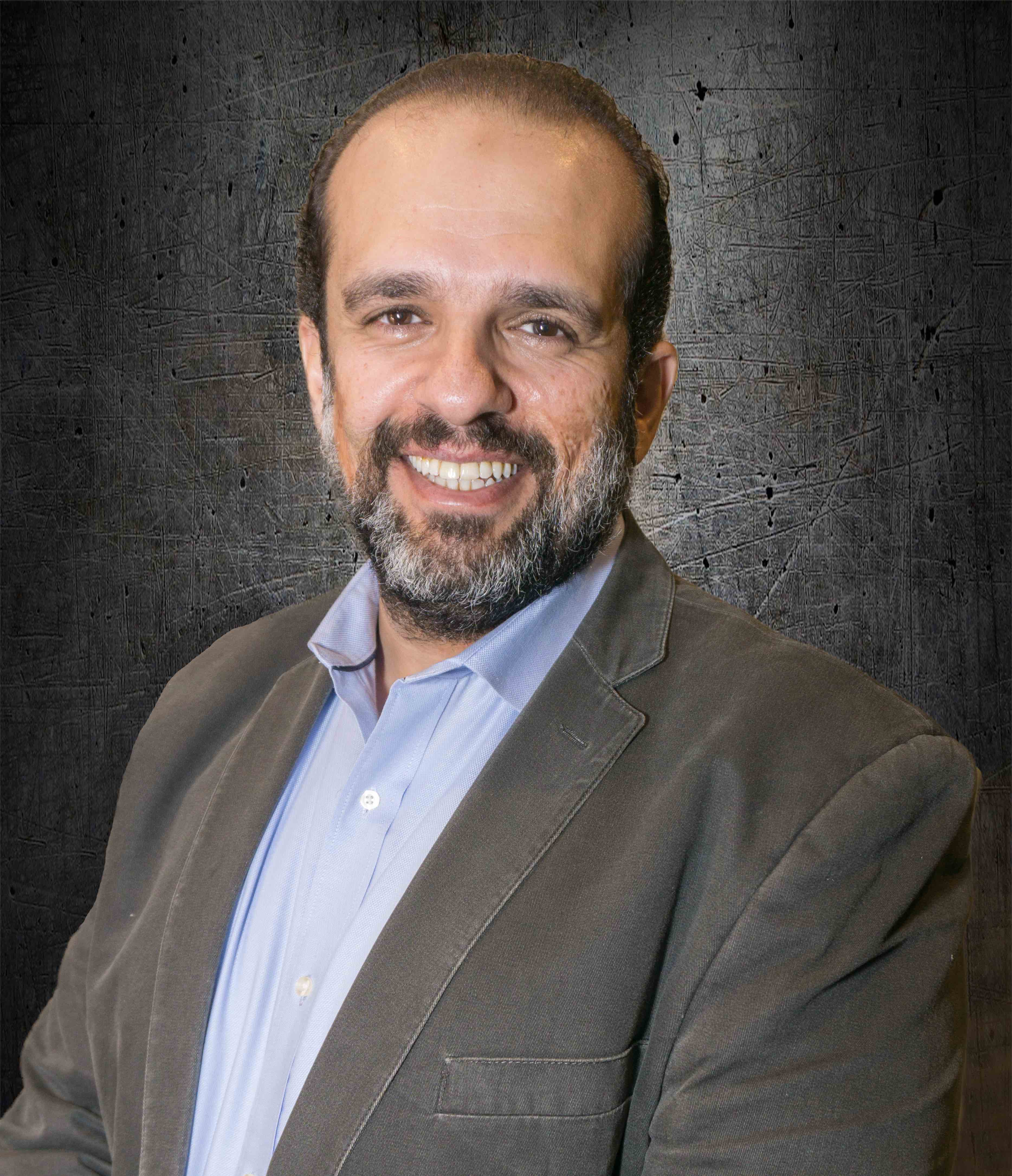}}]{Ghassan AlRegib (SM'22)} is currently the John and Marilu McCarty Chair Professor in the School of Electrical and Computer Engineering, Georgia Institute of Technology. He was a recipient of the ECE Outstanding Graduate Teaching Award in 2001 and both the CSIP Research and the CSIP Service Awards in 2003, the ECE Outstanding Junior Faculty Member Award, in 2008, and the 2017 Denning Faculty Award for Global Engagement.
His research group, the Omni Lab for Intelligent Visual Engineering and Science (OLIVES) works on research projects related to machine learning, image and video processing, image and video understanding, seismic interpretation, healthcare intelligence, machine learning for ophthalmology, and video analytics. He participated in several service activities within the IEEE including the organization of the First IEEE VIP Cup (2017), Area Editor for the IEEE Signal Processing Magazine, and the Technical Program Chair of GlobalSIP14 and ICIP20. He has provided services and consultation to several firms, companies, and international educational and Research and Development organizations. He has been a witness expert in a number of patents infringement cases.
\end{IEEEbiography}



\end{document}